\documentclass[11pt]{article}

\usepackage[final]{acl}
\makeatletter
\acl@anonymizefalse
\makeatother

\usepackage{times}
\usepackage{latexsym}
\usepackage{xcolor}

\usepackage[T1]{fontenc}
\usepackage[utf8]{inputenc}
\usepackage{microtype}
\usepackage{inconsolata}

\usepackage{graphicx}
\usepackage{enumitem}
\usepackage{multirow}
\usepackage{booktabs}
\usepackage{tabularx}
\usepackage{array}
\usepackage{subcaption}
\usepackage[table]{xcolor}

\newcolumntype{Y}{>{\raggedright\arraybackslash}X}
\newcolumntype{F}{>{\ttfamily\raggedright\arraybackslash}p{0.34\textwidth}}

\title{\textsc{MineTRACE}: An Evidence-Grounded Interactive Reasoning System for Mineral Prospectivity}

\author{
  Yiran Zhang\textsuperscript{1}\thanks{Equal contribution}, 
  Jinwen Liu\textsuperscript{1}\footnotemark[1], 
  Daniel Su\textsuperscript{1}\footnotemark[1],
  Yisu Chen\textsuperscript{2},
  Qiang Sun\textsuperscript{1},
  Chris Gonzalez\textsuperscript{1},\\
  \textbf{Eun-Jung Holden\textsuperscript{1}},
  \textbf{Marco Fiorentini\textsuperscript{1}},
  \textbf{Wei Liu\textsuperscript{1}},
  \textbf{Yihao Ding\textsuperscript{1}} \\
  \textsuperscript{1}The University of Western Australia
  \textsuperscript{2}Wilfrid Laurier University \\
  \texttt{yiran.zhang@research.uwa.edu.au, \{wei.liu,yihao.ding\}@uwa.edu.au}\\
}

\begin{document}
\maketitle

\begin{abstract}
Mineral exploration requires integrating heterogeneous geochemical, geophysical, and geological evidence, yet existing prospectivity systems often provide only opaque scores or heatmaps. We present \textsc{MineTRACE}, a web-based system for evidence-grounded exploration of eight commodities: Cu, Au, Ni, W, Sn, Co, Ta, and Mn. Users can explore prospectivity maps, query locations or regions, inspect supporting evidence, and interact through natural language. A transparent expert tree, informed by geological knowledge and known deposits, combines multi-source evidence into interpretable prospectivity scores. For a new location, the conversational assistant retrieves the score and supporting evidence from the analysis pipeline and presents them in natural language. The scorer achieves spatial AUC values of up to 0.917 across different test scenarios, while end-to-end evaluation assesses query accuracy and response grounding. \textsc{MineTRACE} makes public geoscience data easier to access, interpret, and verify, supporting more efficient and transparent mineral exploration. Our video is available via \url{https://geo.nlp-tlp.org/video}, live demo is available via \url{https://geo.nlp-tlp.org}.
\end{abstract}

\section{Introduction}

\begin{figure*}[t]
  \centering
  \includegraphics[width=\linewidth]{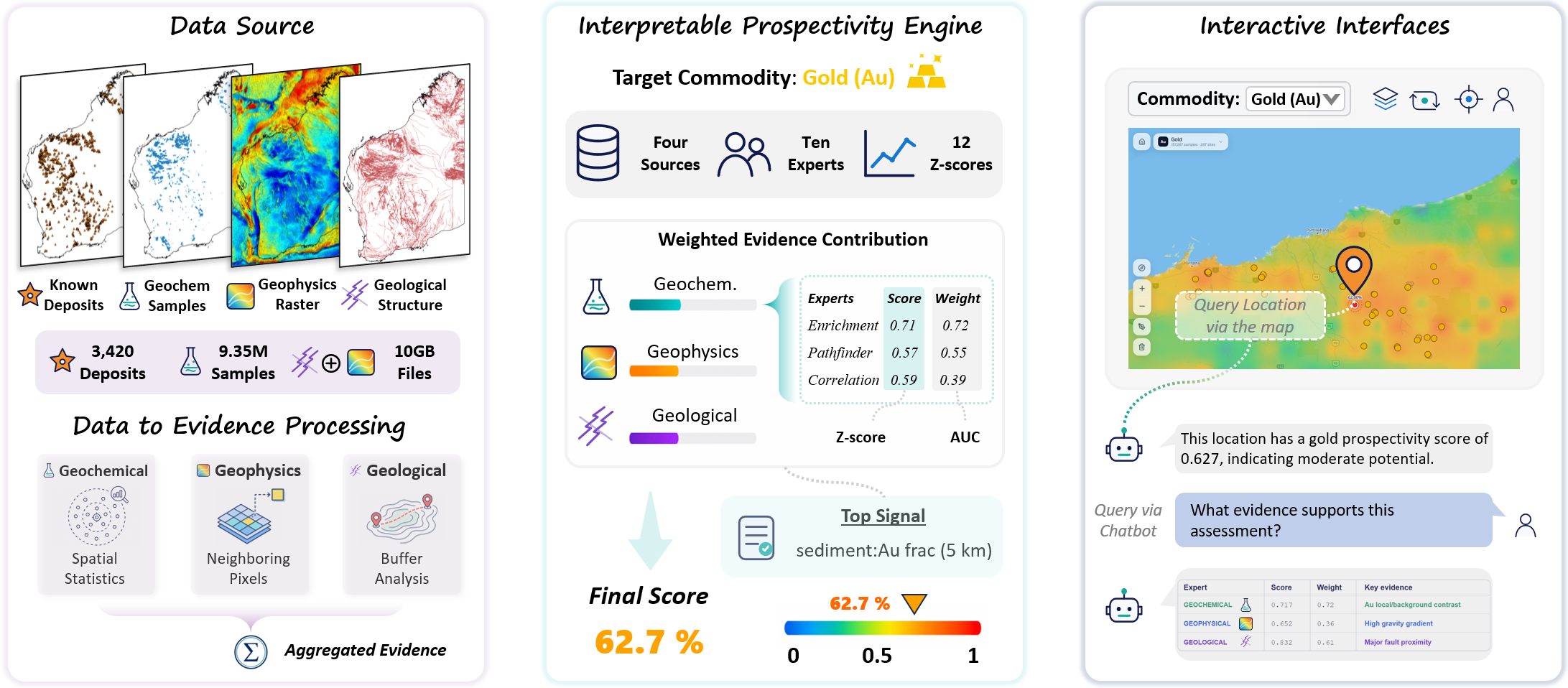}
  \caption{Overview of \textsc{MineTRACE}. Public geochemical, geophysical, and geological evidence is integrated by a transparent expert-tree scorer to assess mineral prospectivity at unknown locations and regions. The scorer returns a unified evidence record that powers the interactive map workspace and grounded natural-language assistant, helping users inspect prospective areas and make evidence-informed exploration decisions.}
  \label{fig:hero}
\end{figure*}

Mineral exploration is essential for expanding the resource base needed to secure critical mineral supplies for modern technologies and economic development \citep{muller2025critical}. Identifying prospective targets, however, requires geologists to integrate heterogeneous geochemical, geophysical, and geological evidence across large and unevenly sampled regions. Mineral prospectivity mapping helps organise this evidence into spatial scores or heatmaps, but practical exploration decisions require more than knowing where a score is high. Geologists must also understand which evidence supports the assessment, whether different evidence sources agree, and whether the available data are sufficient to justify further investigation.

Existing approaches support only parts of this process. Data-driven models can produce accurate prospectivity maps \cite{ding2026geochemad}, but often hide the evidence behind a final score. Knowledge-driven methods are more transparent \cite{dong2024deep}, yet are commonly presented as static maps or offline analyses. Commercial platforms allow users to view multiple data layers, while conversational assistants can simplify interaction \cite{liu2026prisma}, but these components are rarely connected. As a result, users still lack a unified workflow that links each prospectivity assessment to its supporting evidence and allows the result to be inspected interactively within the same interface.

We present \textsc{MineTRACE}, a user-friendly, evidence-centric system designed to bridge the gap between mineral prospectivity predictions and practical exploration decisions. Our aim is to move beyond static scores and heatmaps by making prospectivity assessments interactive, interpretable, and traceable. To support practical exploration workflows, the system allows users to directly query new locations or regions, inspect the multi-source evidence behind each assessment, compare prospective targets, and obtain evidence-grounded explanations through natural language.

\textsc{MineTRACE} supports eight commodities and is built on large-scale public geochemical, geophysical, and geological data from Western Australia. At its core, a transparent expert tree combines nearby observations into a structured evidence record containing the prospectivity score, contributing signals, expert contributions, evidence sources, and local sample coverage. The same record is shared across the map, evidence panels, spatial queries, and conversational interface, ensuring that all system outputs remain consistent with the underlying prospectivity analysis. The contributions of this
paper are summarized as follows:

\begin{itemize}[noitemsep,leftmargin=*]
    
\item \textbf{A user-friendly and explainable system for mineral prospectivity analysis.} \textsc{MineTRACE} integrates exploration, point and region queries, target comparison, evidence inspection, and natural-language interaction, allowing users to obtain prospectivity assessments and understand the evidence behind them within a workflow.

\item \textbf{A traceable multi-source evidence representation.} \textsc{MineTRACE} uses a transparent expert tree to integrate geochemical, geophysical, and geological observations into a structured evidence record, explicitly linking each prospectivity score to its contributing signals, expert contributions, source data, and local coverage.

\item \textbf{A large-scale evaluation of prospectivity and interaction quality.} We evaluate \textsc{MineTRACE} on approximately 9.35 million assays and 3{,}420 known mineral sites across eight commodities, while also assessing prospectivity ranking, spatial generalisation, multi-source evidence fusion, query accuracy, and response grounding.

\end{itemize}

\section{Related Work}
Data-driven approaches model mineral prospectivity as an anomaly detection problem, using methods such as autoencoder–GMMs \cite{wang2025unsupervised} and transformer-based models \cite{yu2026expectation} to capture complex geochemical patterns and achieve strong predictive performance; however, they produce only a single score per location and lack interpretability of the underlying evidence contributing to that score. 
Knowledge-driven approaches instead aim to improve interpretability through structured, tree-based models \cite{zhang2024interpretable, rai2026gold, dong2024deep} that provide feature importance and decision rules for their predictions, but these methods are typically limited to a single data modality, do not generalise well across regions, lack explicit reasoning over multi-source evidence, and are not designed as interactive systems.
Beyond geoscience, recent interactive systems have demonstrated how domain-specific information can be extracted and inspected in a more transparent manner, including key information extraction from domain-specific documents \cite{ding2025synjac}, context-aware spatiotemporal information extraction \cite{zhang2026stindex}, and visual inspection of multi-turn LLM reasoning \cite{zhang2026beyond}, but none of these systems specifically targets mineral prospectivity assessment.
Commercial platforms such as MINML\footnote{\url{https://minml.co.uk/}} attempt to integrate multi-source geoscience data and provide interactive interfaces with prospectivity rankings, claiming evidence supported outputs, without publicly available quantitative evaluation. 
Our MINETRACE provides a unified, evidence-grounded system that links multi-source data, interpretable scoring, and interactive natural-language exploration, with all outputs traceable to underlying evidence and quantitatively evaluated.

\section{System Design}
\label{sec:system}
This section presents the design of \textsc{MineTRACE}. We first introduce the overall system architecture, followed by the data and evidence layer, the interpretable prospectivity engine, and the interactive interfaces and their integration with the system.

\subsection{System Overview}

\label{sec:architecture}

\begin{figure}[t]
  \centering
  \includegraphics[width=\linewidth]{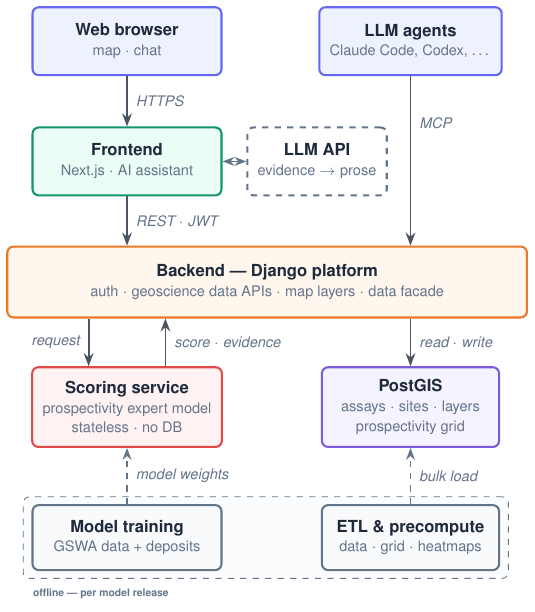}
  \caption{Architecture design for \textsc{MineTRACE}. Django is a Python web framework; PostGIS is a database for geospatial data; ETL means extract--transform--load for data preparation; MCP (Model Context Protocol) lets external agents call software tools.}
  \label{fig:arch}
\end{figure}

\textsc{MineTRACE} is organized as a web system around a single scoring path (Figure~\ref{fig:arch}). Users interact with a browser workspace that combines the map, evidence panel, and conversational assistant. User requests are routed through a platform backend to a separate scoring service, which applies the expert-tree model and returns a prospectivity score together with its evidence record. The scoring service reads processed geoscience data from a PostGIS store and loads fitted expert weights produced by offline training and preprocessing jobs. This design keeps the interface simple while ensuring that every map result and assistant response is generated from the same scoring path.

\subsection{Data and Evidence Layer}
\label{sec:evidence}

\paragraph{Data Sources.}\textsc{MineTRACE} uses three open data products published by the Geological Survey of Western Australia (GSWA). \emph{CM02 Near Surface Geochemistry (Geochem.)} provides 9.35 million assays from five sampling media: sediment, rock chip, drill-hole maximum grade, shallow drilling, and surface soil. \emph{CM01 Mineralization Sites (Sites)} provides 3,420 known mine and deposit sites across eight commodities, which we use to construct supervision labels. The \emph{CM08 Critical Minerals Basemap (Basemap)} provides seven geophysical rasters and five geological vector layers, including magnetics, gravity, radiometrics, geochronology, faults, geological units, and Cenozoic cover. All products are registered to GDA2020 and cover about $2.5$ million~km\textsuperscript{2} across Western Australia.\footnote{Available through the DMIRS Data and Software Centre: \url{https://dasc.dmirs.wa.gov.au}.}

\paragraph{Data to Evidence Processing.} Raw data are first harmonised into a unified statewide evidence store. Geochemical assays are cleaned, mapped to a fixed element schema, and separated by sampling medium; geophysical rasters are aligned to a common spatial reference; and geological, structural, and known-deposit layers are spatially indexed and converted into queryable attributes and distance features. For each query location or region, the system retrieves nearby observations from this store and assembles a structured \emph{evidence record} containing multi-scale geochemical statistics, geophysical values, geological and structural context, proximity to known deposits, and local sample coverage. Missing evidence is handled explicitly: when fewer than three assay samples are available within 10\,km, the geochemical component abstains rather than extrapolating from insufficient observations. This record provides a shared evidence representation for downstream scoring, visual inspection, and direct conversational interaction. Full preprocessing details are provided in Appendix~\ref{app:data}.

\begin{figure*}[t]
    \centering
    \begin{subfigure}[t]{0.63\linewidth}
        \centering
        \includegraphics[width=\linewidth]{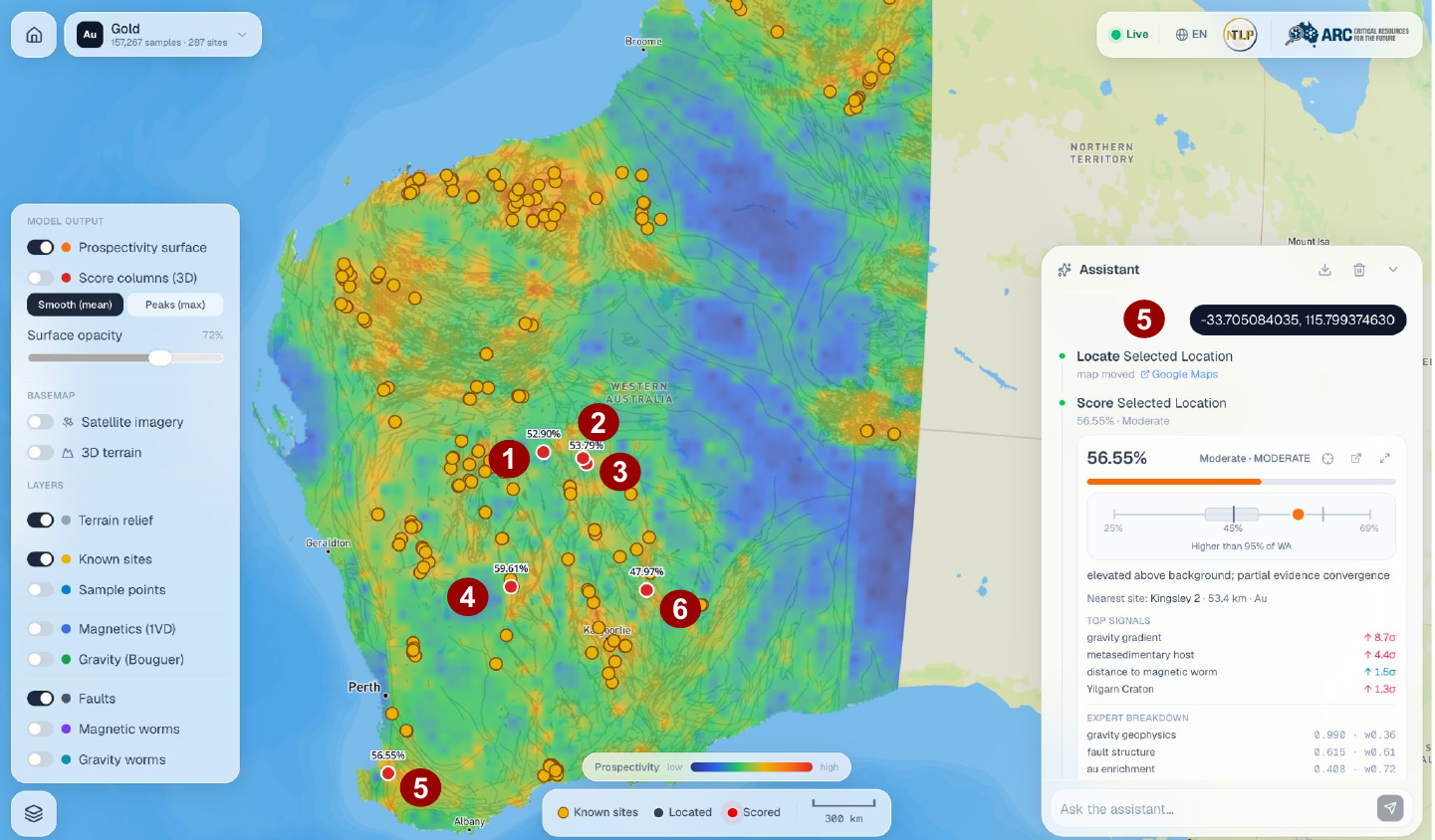}
        \caption{Interactive interface and post-training gold projects.}
        \label{fig:interface}
    \end{subfigure}
    \hfill
    \begin{subfigure}[t]{0.36\linewidth}
        \centering
        \includegraphics[width=\linewidth]{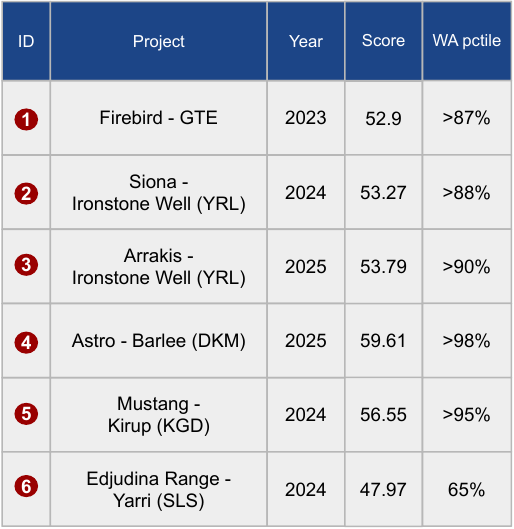}
        \caption{Post-training gold project details.}
        \label{fig:case_study}
    \end{subfigure}

    \caption{Interactive interface and post-training gold-project case study in \textsc{MineTRACE}. \textbf{(a)} The interface illustrates the interaction modes in Section~\ref{sec:interaction}: commodity switching, evidence-layer toggling, map-click prospectivity queries, evidence inspection, and natural-language interaction with the assistant. The six numbered points mark gold projects that became public after the model's training cutoff. \textbf{(b)} Project summaries for these six points. As discussed in Section~\ref{sec:case}, they are used as a qualitative case study rather than a formal benchmark: they were not training positives, and their scores and statewide percentiles show how \textsc{MineTRACE} assesses recent gold targets.}
    \label{fig:interface_case_study}
\end{figure*}

\subsection{Interpretable Prospectivity Engine}
\label{sec:engine}

The interpretable prospectivity engine systematically converts evidence around a query location into a prospectivity score and a traceable, auditable explanation. It has three stages: evidence representation, expert scoring, and score aggregation.

\paragraph{Evidence Representation.}
For each query point, the spatial index retrieves nearby samples within 5, 10, and 50,km. For a fixed panel of 14 target and pathfinder elements, it computes robust statistics in log space, including local enrichment, high-percentile anomalies, local-to-regional contrast, element ratios, spatial coherence, and sample coverage. Geological and geophysical context is added through raster values, mapped attributes, and distances to relevant structures. Locations with insufficient nearby observations are marked as low coverage rather than treated as confident predictions. The detailed feature catalogue at Appendix~\ref{app:features}.

\paragraph{Expert scoring.}
The resulting features are evaluated by named experts, each representing a fixed exploration heuristic and producing a score in $[0,1]$. This design preserves the meaning of each evidence source and allows the final score to be attributed to clear geological arguments.
\textbf{Geochemical experts} evaluate target enrichment, pathfinder enrichment, and correlated element anomalies.
\textbf{Geophysical experts} evaluate magnetic, gravity, radiometric, and geochronological responses.
\textbf{Geological experts} evaluate fault proximity, geophysical worms, favourable host lithologies, and Cenozoic-cover context.
Commodity-specific pathfinder lists and expert definitions are given in Appendix~\ref{app:experts}.

\paragraph{Transparent Score aggregation.}
The final prospectivity score is a weighted mean of the active expert scores. Expert weights are fitted offline according to how well each expert separates known deposits from background locations. Experts without the required evidence abstain and are excluded from the aggregation, while a separate coverage value indicates how much local evidence supports the score.
No end-to-end neural model is trained. Offline fitting estimates expert and source weights, background statistics for z-scoring, and the favourable direction of each feature. Pathfinder definitions remain fixed from domain knowledge.
Each prediction is stored as an evidence record containing the contributing features, z-scores, expert scores, expert weights, active sources, coverage information, and final commodity score. The fitted parameters are frozen at deployment, and the deployed scorer matches the reference implementation to within $10^{-6}$ on a fixed golden set. This ensures that offline evaluation and interactive queries use the same audited scoring process.

\subsection{Interactive Interfaces}
\label{sec:interaction}

The interface exposes the same evidence record through three complementary interaction modes.

\paragraph{Direct spatial querying.}
A user selects a commodity and scores any location by clicking a point or drawing a region. The system returns a prospectivity value, a tier, local sample coverage, nearest known-deposit context, and an evidence panel. The panel renders the model's own decomposition: ranked feature signals, z-scores, source medium, contributing experts, expert weights, and active evidence layers. This lets the user inspect why the location scores as it does.

\paragraph{Contextual map exploration.}
The workspace serves a precomputed prospectivity surface for each commodity as a toggleable heatmap. Known deposits, magnetic and gravity layers, faults, worms, and other context layers can be displayed alongside the score surface. Away from local data, the surface is marked as interpolated rather than treated as direct evidence. This helps users read a target in geological context and compare a geochemical anomaly with independent evidence.

\paragraph{Conversational tool use.}
A user can also ask in natural language. The assistant scores and compares locations, ranks regions for a metal, switches the active commodity, moves the map, and toggles evidence layers by calling system tools. Because it answers from the same evidence records shown in the evidence panels, the assistant is a queryable interface to the interpretable model rather than an independent narrator. In particular, since LLM assistants can otherwise drift toward plausible but unsupported claims \cite{kashyap2026thinkalignllmshelpful}, it is instructed to report only tool-computed scores, distances, tiers, and rankings.

\section{Evaluation}
\label{sec:eval}
\subsection{Scoring Quality}

\paragraph{Evaluation Setup.}
We evaluate whether the scorer ranks known mineralisation above background across eight commodities. Positive samples are sites classified as \emph{Mine} or \emph{Deposit}. We consider four negative-sampling strategies of increasing difficulty: random locations across Western Australia, far-random locations more than 50\,km from any known site, validated sites associated with other commodities (\emph{NonMine}), and a south-train/north-test spatial holdout (\emph{Spatial}). For each setting, we hold out 30 positive sites, sample 200 negatives, fit the expert weights on the remaining data, and report ROC-AUC. Because random and far-random negatives are often located in sparsely sampled areas, we treat the NonMine and Spatial settings as the more realistic tests.

\begin{table}[h]
\centering
\footnotesize
\setlength{\tabcolsep}{5.0pt}
\renewcommand{\arraystretch}{1.12}
\begin{tabular}{llcclcc}
\hline
 &  & \multicolumn{2}{c}{\cellcolor[HTML]{EEF2F6}\textbf{Context tests}} &  & \multicolumn{2}{c}{\cellcolor[HTML]{E8F4F6}\textbf{Conservative tests}} \\ \cline{3-4} \cline{6-7} 
\multirow{-2}{*}{\textbf{Metal}} &  & \cellcolor[HTML]{EEF2F6}\textbf{Random} & \cellcolor[HTML]{EEF2F6}\textbf{Far-R.} &  & \cellcolor[HTML]{E8F4F6}\textbf{NonMine} & \cellcolor[HTML]{E8F4F6}\textbf{Spatial} \\ \hline
Cu &  & 0.861 & 0.925 &  & 0.629 & 0.576 \\
Au &  & 0.925 & 0.958 &  & 0.820 & 0.730 \\
Ni &  & \cellcolor[HTML]{2F6690}{\color[HTML]{FFFFFF} 0.990} & \cellcolor[HTML]{2F6690}{\color[HTML]{FFFFFF} 0.994} &  & \cellcolor[HTML]{2F6690}{\color[HTML]{FFFFFF} 0.946} & \cellcolor[HTML]{2F6690}{\color[HTML]{FFFFFF} 0.917} \\
W &  & 0.907 & 0.936 &  & 0.740 & 0.741 \\
Sn &  & 0.972 & 0.989 &  & 0.895 & 0.664 \\
Co &  & 0.977 & 0.983 &  & 0.929 & 0.805 \\
Ta &  & 0.919 & 0.953 &  & 0.775 & 0.782 \\
Mn &  & 0.802 & 0.825 &  & 0.745 & 0.257 \\ \hline
{\color[HTML]{1F4E5F} \textbf{Average}} & {\color[HTML]{1F4E5F} } & {\color[HTML]{1F4E5F} \textbf{0.919}} & {\color[HTML]{1F4E5F} \textbf{0.945}} & {\color[HTML]{1F4E5F} } & {\color[HTML]{1F4E5F} \textbf{0.810}} & {\color[HTML]{1F4E5F} \textbf{0.684}} \\ \hline
\end{tabular}
  \caption{Prospectivity ranking across eight commodities, measured by
  ROC-AUC. Random and Far-Random are contextual tests, while NonMine and
  Spatial provide more conservative evaluations. Highlighted cells indicate
  the best commodity-level result in each column.}
  \label{tab:auc}
\end{table}

\begin{figure}[h]
  \centering
  \includegraphics[width=\columnwidth]{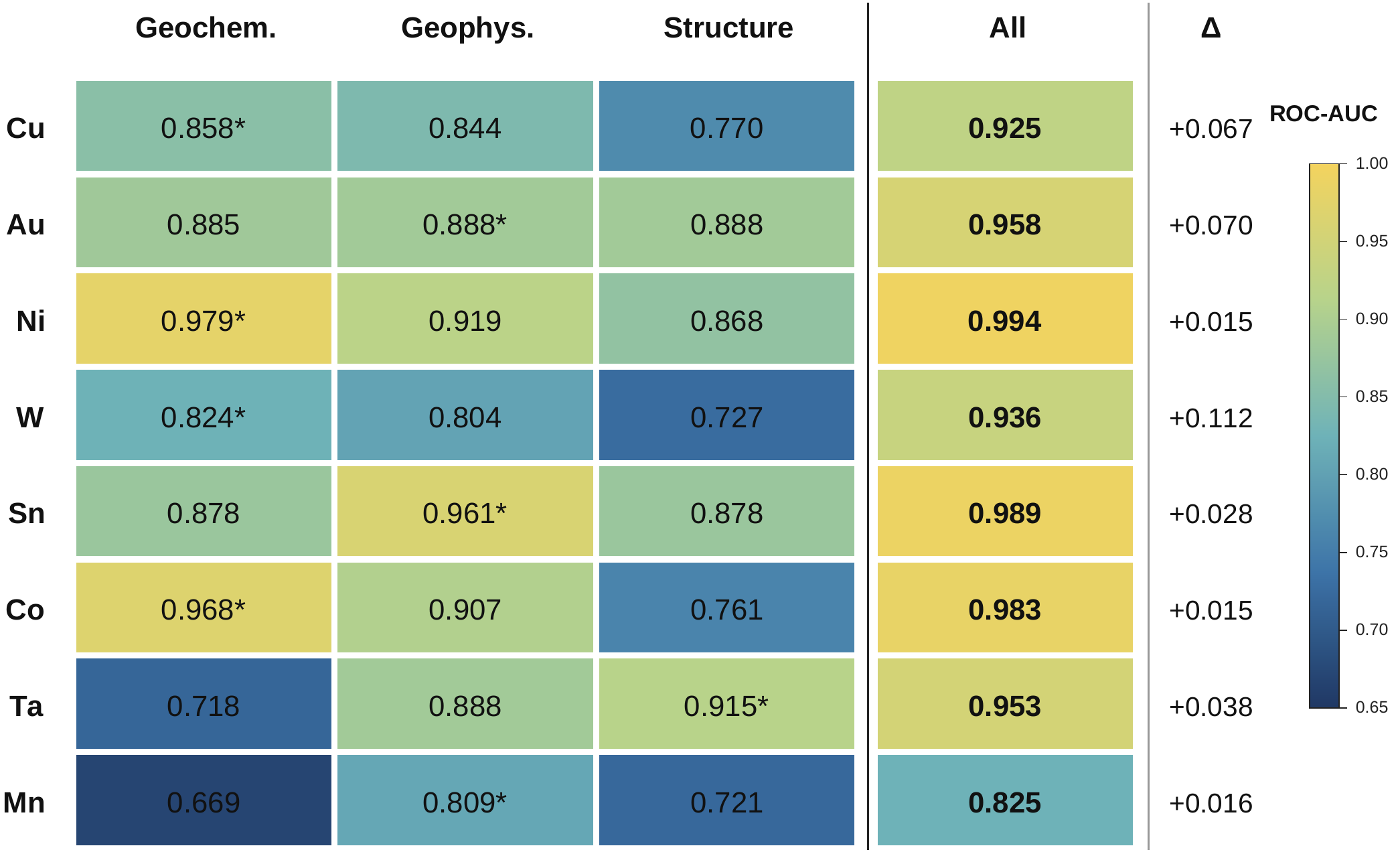}
  \caption{Evidence-family ablation under ROC-AUC; $\Delta$ is the fused model’s gain over the best single family.}
  \label{fig:evidence-fusion}
\end{figure}

\paragraph{Results and insights.}
Table~\ref{tab:auc} shows that performance varies across commodities and test settings. Ni performs best on the realistic tests, reaching 0.946 NonMine AUC and 0.917 Spatial AUC, while Co also remains strong at 0.929 and 0.805. Au and Sn perform well against validated non-target sites, whereas Cu and Mn degrade substantially under spatial holdout, indicating weaker regional generalisation. Figure~\ref{fig:evidence-fusion} further shows that geochemical, geophysical, and geological evidence are each informative, but their importance differs by commodity. Although this ablation uses the easier far-random setting, combining all three evidence families consistently gives the highest AUC for every commodity. These results collectively support both multi-source fusion and the interface design that exposes evidence contributions and local coverage alongside each score.

\subsection{End-to-End Human Evaluation}

\paragraph{Evaluation Setup.}
To simulate realistic use, we designed a suite of 30 natural-language test questions and had five human evaluators assess the deployed system end to end, complementing benchmark-style evaluations of multi-turn reasoning and cross-domain model capability \cite{zhang-etal-2025-turnbench, joshi2026argusjudgeallcomparing}.  Each evaluator independently posed all 30 questions to the live system and rated every response \emph{Good} or \emph{Bad} against a pre-specified expected behaviour, recording a failure category for each \emph{Bad} response. We report end-to-end functional success and cross-trial consistency; the full question list, capability groups, and failure-category definitions are given in Appendix~\ref{app:humaneval}.

\paragraph{Results and insights.}
Across the 150 question--evaluator trials, 92\% of responses were rated Good (83--100\% per evaluator), and consistency is high: 21 of the 30 questions were rated Good by all five evaluators and 29 of 30 by a majority (Table~\ref{tab:human-eval}). Every capability exceeds 87\% except interpolation disclosure (70\%). Grounding is strong, only one of the 150 responses was flagged for a fabricated number, and the dominant failures were map-pin placement and occasional wrong-tool calls rather than unsupported claims. With the scoring evaluation and the blind discovery test (Section~\ref{sec:case}), this shows that \textsc{MineTRACE} performs its functions reliably while keeping its answers grounded.

\begin{table}[h]
\centering
\footnotesize
\setlength{\tabcolsep}{5pt}
\renewcommand{\arraystretch}{1.1}
\begin{tabular}{llcc}
\hline
\textbf{Capability (\# questions)} &  & \textbf{Trials} & \textbf{Success} \\ \hline
Locate \& score (6) &  & 30 & 0.93 \\
Top-prospect ranking (5) &  & 25 & 1.00 \\
Layer control (4) &  & 20 & 0.95 \\
Scope \& grounding (4) &  & 20 & 0.95 \\
Comparison \& recommendation (6) &  & 30 & 0.87 \\
Evidence \& clarification (3) &  & 15 & 0.93 \\
Coverage/interpolation disclosure (2) &  & 10 & 0.70 \\ \hline
\textbf{Overall (30)} &  & \textbf{150} & \textbf{0.92} \\ \hline
\end{tabular}
\caption{End-to-end human evaluation. \emph{Trials} is the number of question--evaluator judgements; \emph{Success} is the fraction rated Good.}
\label{tab:human-eval}
\end{table}

\section{Case Study: A Blind Test on New Gold Discoveries}
\label{sec:case}

The evaluation above uses known deposits. A more demanding question for a prospectivity system is whether it can highlight mineralisation that was still unknown when the model was built. We therefore conduct a retrospective blind case study on six Western Australian gold discoveries first announced by ASX-listed explorers in 2023--2025 \citep{disc:mustang,disc:edjudina,disc:firebird,disc:Barlee,disc:siona,disc:arrakis}. Because \textsc{MineTRACE} was built from public data compiled to 2021--2022 (Appendix~\ref{app:data}), and none of these discoveries appears in the training labels, the case study tests whether the deployed system assigns high prospectivity to targets it could not have memorised.

For each discovery, we use the coordinate reported in the company announcement or the closest available prospect centroid, score that point with the deployed model, and compare the score with the statewide gold prospectivity surface. Figure~\ref{fig:interface_case_study} shows the resulting locations and scores. Five of the six discoveries fall in the top 13\% of Western Australia, three fall in the top 10\%, and the strongest case, Astro at the Barlee project, lies above the 98th percentile. The only mid-ranked site is Edjudina Range, which still scores near the 65th percentile. Because these targets were not part of the supervision set, these results suggest that the model is not merely rediscovering labelled deposits: it concentrates prospectivity in areas where new gold systems were later reported.

We use Mustang as a worked example to show why an evidence-grounded interface is useful (Figure~\ref{fig:interface}). Mustang had no effective prior drilling in the model data, and its nearest catalogued gold deposit is approximately 53 km away. Its score therefore cannot be explained simply by proximity to a known label. The local gold-geochemical evidence is also weak, because the drilling that reported gold mineralisation post-dates the public data used by the system. Instead, the evidence panel shows that the moderate-to-high score is driven by independent geological and geophysical support: a strong gravity-gradient signal, favourable metasedimentary host context, Yilgarn Craton membership, and structural-context features involving gravity worms and mapped faults. These signals are consistent with the company's description of Mustang as an early-stage gold prospect in the Southwest Terrane Greenstones of the southwestern Yilgarn Craton, where mineralisation is associated with a significant shear-zone setting \citep{disc:mustang}. In other words, \textsc{MineTrace} did not ``know'' that Mustang contained gold; it surfaced a defensible mineral-systems argument that a geologist could inspect before the later drilling result was public.

This case study also clarifies the system's intended role.The six discoveries are early-stage exploration results, not defined resources, and the sample is too small for statistical validation. Absolute scores are moderate in several cases, and the high percentile rankings partly reflect the generally low predicted gold prospectivity across Western Australia.  Its value is therefore in the workflow it demonstrates: \textsc{MineTrace} surfaces plausible targets, exposes the evidence behind each score. For exploration decision support, this traceability matters as much as the ranking itself.

\section{Conclusion}

We presented \textsc{MineTRACE}, a user-friendly and evidence-grounded interactive reasoning system for mineral prospectivity. The system turns public exploration data into traceable evidence records that connect final scores to measured features, named experts, expert weights, source layers, and coverage metadata. The same evidence structure powers the map interface, heatmaps, evidence panels, APIs, MCP endpoint, and a grounded conversational assistant. This design turns a prospectivity model from a static score generator into an interactive reasoning workflow. \textsc{MineTRACE} demonstrates that interpretable domain models can support natural-language scientific interaction when their evidence structure is preserved end to end.

\section*{Limitations}
\label{sec:limitations}

\textsc{MineTRACE} is a decision-support system, not an autonomous or field-validated one: it ranks and explains evidence from public data but does not replace field verification or expert geological judgement, and its scores are bounded by data coverage, assay quality, spatial sampling bias, and the completeness of public reporting, such that areas with sparse local samples receive weak or interpolated support that the interface must flag. The expert tree, while transparent, remains a simplified model of mineral systems; its fixed expert set aids auditability but may miss commodity-specific processes, and performance is uneven across commodities, being strong for Ni and Co but weak for Cu and Mn under spatial holdout. Similarly, although the conversational assistant is grounded through tools, grounding is not correctness, since it is only as reliable as the scores and evidence records it retrieves; we verify that numerical claims are tool-derived and that unsupported requests are declined, but larger user studies are still needed to measure how geologists use the system in practice.

\section*{Acknowledgement}

This research was gratefully supported by the Australian Research Council (ARC) Training Centre for Critical Resources for the Future (CCRF) under grant number IC230100035. The authors acknowledge the computational resources and technical support provided by the Kaya High Performance Computing facility at the University of Western Australia (UWA), which were essential to the data processing and analysis conducted in this study. We also thank the Western Australian Mineral Exploration (WAMEX) database, administered by the Geological Survey of Western Australia, for providing free and public access to the exploration report data used throughout this work.

\bibliography{custom}

@article{ding2026geochemad,
  title={GeoChemAD: Benchmarking Unsupervised Geochemical Anomaly Detection for Mineral Exploration},
  author={Ding, Yihao and Zhang, Yiran and Gonzalez, Chris and Holden, Eun-Jung and Liu, Wei},
  journal={arXiv preprint arXiv:2603.13068},
  year={2026}
}

@article{muller2025critical,
  title={Critical metals: Their mineral systems and exploration},
  author={M{\"u}ller, Daniel and Groves, David I and Santosh, M and Yang, Cheng-Xue},
  journal={Geosystems and Geoenvironment},
  volume={4},
  number={1},
  pages={100323},
  year={2025},
  publisher={Elsevier}
}

@article{wang2025unsupervised,
  title={Unsupervised detection of multivariate geochemical anomalies using a high-performance deep autoencoder Gaussian mixture model},
  author={Wang, Xuemei and Chen, Yongliang},
  journal={Journal of Geochemical Exploration},
  volume={271},
  pages={107671},
  year={2025},
  publisher={Elsevier}
}

@article{yu2026expectation,
  title={Expectation--Maximization-Derived Self-distillation Meets Transformer: A Robust Unsupervised Deep Learning Approach for Geochemical Anomaly Recognition},
  author={Yu, Shuyan and Deng, Hao and Liu, Xinyu and Zheng, Yang and Liu, Zhankun and Chen, Jin and Mao, Xiancheng},
  journal={Mathematical Geosciences},
  volume={58},
  number={2},
  pages={279--312},
  year={2026},
  publisher={Springer}
}

@misc{zhang2024interpretable,
  title={Interpretable machine learning for geochemical anomaly delineation in the yuanbo nang district, gansu province, china. Minerals, 14 (5): 500},
  author={Zhang, S and Carranza, EJM and Fu, C and Wen-zhi, Z and Xiang, Q},
  year={2024}
}

@article{rai2026gold,
  title={Gold prospectivity mapping in the eastern part of Mahakoshal Fold Belt, India: A comparative study of random forest and XGBoost leveraging knowledge-guided feature engineering},
  author={Rai, Apratim Kumar and Tripathi, Utkarsh and Kumar, Gautam and Siddique, Shahina and Singh, Vinay and Sathikumar, Resmi and Bagchi, Joyesh},
  journal={Geosystems and Geoenvironment},
  pages={100532},
  year={2026},
  publisher={Elsevier}
}

@article{dong2024deep,
  title={Deep forest modeling: an interpretable deep learning method for mineral prospectivity mapping},
  author={Dong, Yue-Lin and Zhang, Zhen-Jie},
  journal={Journal of Geophysical Research: Machine Learning and Computation},
  volume={1},
  number={4},
  pages={e2024JH000311},
  year={2024},
  publisher={Wiley Online Library}
}

@misc{disc:mustang,
  author       = {{Kula Gold Limited}},
  title        = {Mustang Gold Prospect – Results Update},
  year         = {2025},
  howpublished = {ASX Announcement},
  url          = {https://www.kulagold.com.au/wp-content/uploads/2025/04/02935171.pdf},
  note         = {Accessed: 2026-07-08}
}

@misc{disc:edjudina,
  author       = {{Solstice Minerals Limited}},
  title        = {Edjudina Range Gold Discovery Ready for First RC Drilling},
  year         = {2025},
  month        = may,
  howpublished = {ASX Announcement},
  url          = {https://solsticeminerals.com.au/upload/documents/investor/asx/250502003137_250502EdjudinaRangeGoldDiscoveryReadyForFirstRCDrillingfinal.pdf},
  note         = {Accessed: 2026-07-08}
}

@misc{disc:firebird,
  author       = {{Great Western Exploration Limited}},
  title        = {Firebird Aircore Drilling Results Received},
  year         = {2024},
  month        = feb,
  howpublished = {ASX Announcement},
  url          = {https://company-announcements.afr.com/asx/gte/7e1ccabc-cab5-11ee-be79-0abdb9403284.pdf},
  note         = {Accessed: 2026-07-08}
}

@misc{disc:Barlee,
  author       = {{Duketon Mining Limited}},
  title        = {September 2024 Quarterly Report},
  year         = {2024},
  month        = oct,
  howpublished = {ASX Announcement},
  url          = {https://announcements.asx.com.au/asxpdf/20241024/pdf/069hcpnnf7mvhq.pdf},
  note         = {Accessed: 2026-07-08}
}

@misc{disc:siona,
  author       = {{Yandal Resources Ltd}},
  title        = {Emerging Gold Discovery within the New England Granite Prospect},
  year         = {2024},
  month        = oct,
  howpublished = {ASX Announcement},
  url          ={https://announcements.asx.com.au/asxpdf/20241021/pdf/069bssr29yf7qq.pdf},
  note         = {Accessed: 2026-07-08}
}

@misc{disc:arrakis,
  author       = {{Yandal Resources Ltd}},
  title        = {Arrakis Gold Discovery Confirmed With 54 m @ 1.2 g/t Au from 108 m},
  year         = {2025},
  month        = sep,
  howpublished = {ASX Announcement},
  url          = {https://announcements.asx.com.au/asxpdf/20250922/pdf/06pgs3znt24cpy.pdf},
  note         = {Accessed: 2026-07-08}
}

@article{ding2025synjac,
  title={Synjac: Synthetic-data-driven joint-granular adaptation and calibration for domain specific scanned document key information extraction},
  author={Ding, Yihao and Han, Soyeon Caren and Li, Zechuan and Chung, Hyunsuk},
  journal={Information Fusion},
  pages={104074},
  year={2025},
  publisher={Elsevier}
}

@inproceedings{zhang2026stindex,
  title={STIndex: A Context-Aware Multi-Dimensional Spatiotemporal Information Extraction System},
  author={Zhang, Wenxiao and Liu, Yu and Sun, Qiang and Ding, Yihao and Li, Sirui and Liu, Yanbing and Hong, Jin B and Liu, Wei},
  booktitle={Companion Proceedings of the ACM Web Conference 2026},
  pages={69--72},
  year={2026}
}

@article{liu2026prisma,
  title={PRISMA: Reinforcement Learning Guided Two-Stage Policy Optimization in Multi-Agent Architecture for Open-Domain Multi-Hop Question Answering},
  author={Liu, Yu and Zhang, Wenxiao and Cao, Cong and Lu, Wenxuan and Yuan, Fangfang and Guo, Diandian and Peng, Kun and Sun, Qiang and Zhang, Kaiyan and Liu, Yanbing and others},
  journal={arXiv preprint arXiv:2601.05465},
  year={2026}
}

@misc{joshi2026argusjudgeallcomparing,
      title={Can Argus Judge Them All? Comparing VLMs Across Domains}, 
      author={Harsh Joshi and Gautam Siddharth Kashyap and Rafiq Ali and Ebad Shabbir and Niharika Jain and Sarthak Jain and Jiechao Gao and Usman Naseem},
      year={2026},
      eprint={2507.01042},
      archivePrefix={arXiv},
      primaryClass={cs.IR},
      url={https://arxiv.org/abs/2507.01042}, 
}

@misc{kashyap2026thinkalignllmshelpful,
      title={We Think, Therefore We Align LLMs to Helpful, Harmless and Honest Before They Go Wrong}, 
      author={Gautam Siddharth Kashyap and Mark Dras and Usman Naseem},
      year={2026},
      eprint={2509.22510},
      archivePrefix={arXiv},
      primaryClass={cs.CL},
      url={https://arxiv.org/abs/2509.22510}, 
}

@inproceedings{zhang-etal-2025-turnbench,
    title = "{T}urn{B}ench-{MS}: A Benchmark for Evaluating Multi-Turn, Multi-Step Reasoning in Large Language Models",
    author = "Zhang, Yiran  and
      Wang, Mo  and
      Li, Xiaoyang  and
      Ren, Kaixuan  and
      Zhu, Chencheng  and
      Naseem, Usman",
    editor = "Christodoulopoulos, Christos  and
      Chakraborty, Tanmoy  and
      Rose, Carolyn  and
      Peng, Violet",
    booktitle = "Findings of the Association for Computational Linguistics: EMNLP 2025",
    month = nov,
    year = "2025",
    address = "Suzhou, China",
    publisher = "Association for Computational Linguistics",
    url = "https://aclanthology.org/2025.findings-emnlp.1084/",
    doi = "10.18653/v1/2025.findings-emnlp.1084",
    pages = "19892--19924",
    ISBN = "979-8-89176-335-7",
}

@inproceedings{zhang2026beyond,
  title={Beyond the black box: Demystifying multi-turn llm reasoning with vista},
  author={Zhang, Yiran and Lin, Mingyang and Dras, Mark and Naseem, Usman},
  booktitle={Proceedings of the AAAI Conference on Artificial Intelligence},
  volume={40},
  number={48},
  pages={41745--41747},
  year={2026}
}

\newpage
\appendix

\section{Data and Preprocessing Details}
\label{app:data}

The corpus comprises 9{,}352{,}545 assay samples across five sampling media (Table~\ref{tab:data-media}) and 3{,}420 positive sites across eight commodities: 1{,}057~Cu, 581~Ni, 492~Sn, 383~Co, 306~Ta, 287~Au, 215~Mn, and 99~W. Geochemistry is drawn from GSWA \emph{CM02 Near Surface Geochemistry}, deposit labels from \emph{Mineralization Sites}, and the geophysical rasters and geological vectors from the 2021 \emph{CM08 Critical Minerals Basemap}; all products are openly licensed and standardised to GDA2020. 

For preprocessing, we map all \textit{CM02 Geochem.} records to a common schema containing coordinates, sampling medium, and 123 element or oxide fields. We convert the $-9999$ sentinel to missing, set negative below-detection values to zero, split the samples into five medium-specific tables, and apply the transformation $\log(1{+}x)$ to reduce skew. For CM01, we merge the commodity-specific exports into one supervision table per target commodity, treating polymetallic sites as positive for each associated commodity. We retain sites classified as \emph{Mine} or \emph{Deposit} and exclude \emph{Prospect} and \emph{Occurrence} records. All processed assay, site, raster, and vector layers are stored in PostGIS. When no assay samples fall within the query radius, the system returns no geochemical score rather than extrapolating into unsupported areas. 

\begin{table}[h]
\centering
\footnotesize
\begin{tabular}{lllllr}
\hline
\textbf{Sampling medium} & & & & & \textbf{Samples} \\ \hline
Stream sediment & & & & & 157{,}267 \\
Rock chip & & & & & 402{,}283 \\
Drill hole (max grade) & & & & & 1{,}836{,}002 \\
Shallow drill hole & & & & & 1{,}549{,}340 \\
Surface soil & & & & & 5{,}407{,}653 \\ \hline
Total & & & & & 9{,}352{,}545 \\ \hline
\end{tabular}
\caption{Per-medium assay-sample counts (GSWA \emph{CM02 Near Surface Geochemistry}).}
\label{tab:data-media}
\end{table}

\section{Neighbourhood Feature Catalogue}
\label{app:features}

Every feature is a statistic over samples within a neighbourhood radius $s \in \{5, 10, 50\}$ km of the query location, computed on $\log(1{+}x)$ concentrations to reduce the effect of heavy censoring in assay data. Element-level features are instantiated for each element in a fixed 14-element panel (target metals and common pathfinders); target metals outside this panel (Ta, Mn) therefore have no element-level features for the target element itself. Detailed feature descriptions are available at \url{https://github.com/grantzyr/GeoResearchPlatformPublic}. Section~\ref{sec:engine} describes how the model composes them into a score.

\section{Pathfinders and Expert Definitions}
\label{app:experts}

\paragraph{Commodity pathfinders.}
Each commodity is scored on its target element and a fixed suite of pathfinder elements and element ratios drawn from exploration geochemistry. The pathfinder weights encode how diagnostic each element is of the target system and are fixed from domain knowledge, whereas the favourable direction of every feature is learned from data during offline fitting. Target metals outside the 14-element panel (Ta, Mn) are scored through these pathfinders and ratios rather than through a target-element anomaly.

\paragraph{Expert definitions.}
The ten experts are organised into three families (Table~\ref{tab:experts}). The three geochemical experts each aggregate evidence across all active assay media (Table~\ref{tab:data-media}): every medium is fitted independently and weighted by its training ROC-AUC, so the geochemical expert count is fixed at three regardless of how many media cover a query. The four geophysical and three geological experts are single-source leaves over the corresponding raster and vector layers. Within every expert, features are z-scored against a fitted background, combined by a weighted mean under a learned favourable direction ($\pm1$) per feature, and squashed to $[0,1]$; an expert whose required evidence is absent abstains and is dropped from the aggregation.

\begin{table*}[t]
  \centering
  \footnotesize
  \renewcommand{\arraystretch}{1.3}
  \begin{tabularx}{\textwidth}{@{}l l Y@{}}
    \toprule
    \textbf{Family} & \textbf{Expert} & \textbf{Evidence evaluated and primary inputs} \\
    \midrule
    \multirow{6}{*}{Geochemical}
      & Target enrichment   & Local enrichment of the target element above regional background: local-to-regional log contrasts (5--50\,km, 10--50\,km) and high-percentile and fraction-above statistics at 5 and 10\,km. \\
      & Pathfinder          & Anomalies in the commodity's pathfinder elements and element ratios, weighted by the domain pathfinder weights. \\
      & Element correlation & Joint co-enrichment of element pairs among the target and its pathfinders, scored as the product of their local contrasts. \\
    \midrule
    \multirow{4}{*}{Geophysical}
      & Magnetic       & Magnetic intensity and gradient. \\
      & Gravity        & Bouguer gravity and gradient. \\
      & Radiometric    & K, Th, and U channels, their gradients, and the K/Th, Th/U, and U/K ratios. \\
      & Geochronology  & Isotopic crustal-age proxies (Lu--Hf, Sm--Nd) and their gradients. \\
    \midrule
    \multirow{4}{*}{Geological}
      & Fault    & Distance to the nearest mapped fault and fault density at 5 and 10\,km. \\
      & Worm     & Distance to magnetic and gravity worms (multi-scale potential-field edges). \\
      & Geology  & Host-rock lithology class (granitic, felsic, mafic, ultramafic, metasedimentary, sedimentary, metamorphic, hydrothermal), crustal age, craton membership (Yilgarn, Pilbara), and Cenozoic-cover flag. \\
    \bottomrule
  \end{tabularx}
  \caption{The ten named experts across three families. Geochemical experts aggregate over all active assay media; geophysical and geological experts are single-source leaves over the geophysical rasters and geological vector layers, respectively.}
  \label{tab:experts}
\end{table*}

\section{Human Evaluation Details}
\label{app:humaneval}

The 30 questions used in Section~\ref{sec:eval} exercise seven user-facing capabilities, each targeting a distinct part of the exploration workflow (Table~\ref{tab:eval-questions}). Each \emph{Bad} response was tagged with one of ten failure categories: fabricated data or score; coverage/interpolation not disclosed; map-location or pin error; layer-toggle error; wrong tool call or workflow not followed; scope-handling error (wrong refusal or acceptance); unsound comparison or recommendation; unclear evidence explanation; ambiguity not clarified; and other.

\begin{table*}[t]
\centering
\footnotesize
\begin{tabularx}{\linewidth}{@{}p{0.24\linewidth}X@{}}
\toprule
\textbf{Capability} & \textbf{Example questions} \\
\midrule

Locate \& score &
What's the copper potential near Kalgoorlie?;
Score the point 121.47, $-30.75$ for nickel;
How prospective is postcode 6430 for copper?;
Look at Perth---how prospective is it for copper, and what's nearby?;
Zoom to the Pilbara region;
Find prospective copper around the Fortescue area and show it on the map. \\

Top-prospect ranking &
Where in WA is most prospective for gold?;
Switch to tungsten and show me where it's most prospective;
Give me three separate high-potential gold areas, not clustered together;
Score a good gold spot for me---you choose;
Rank the best manganese areas and drop pins so I can see them. \\

Layer control &
Show me the magnetics and faults layers;
Turn off the prospectivity surface and just show known sites;
What geophysical layers can help me interpret a copper target?;
Show gravity worms and magnetic worms together. \\

Scope \& grounding &
Is there any mineral potential in Sydney?;
What about rare earth elements around Mount Weld?;
Which metals can this model actually score?;
Score Newman for iron ore. \\

Comparison \& recommendation &
Compare copper prospectivity between Kalgoorlie and Leonora;
Find the top 5 cobalt targets and tell me which is the best bet and why;
Which is more prospective for nickel: Kambalda or Ravensthorpe?;
Compare the top tin prospect with the top tantalum prospect;
Is Kalgoorlie better for gold or nickel?;
For copper, where should a junior explorer focus in WA, and why? \\

Evidence \& clarification &
I've selected a point on the map---how is it for copper?;
Why is this location scored high---walk me through the signals;
I want a town starting with `Mee'---can you help me pick? \\

Coverage/interpolation disclosure &
Is the score at Telfer based on real data there, or interpolated?;
What's the prospectivity at $-25.0$, 122.0 (a remote, likely uncovered point)? \\

\bottomrule
\end{tabularx}
\caption{The 30 human-evaluation questions grouped by capability (cf.\ Table~\ref{tab:human-eval}).}
\label{tab:eval-questions}
\end{table*}

\end{document}